\documentclass[10pt, onecolumn]{article}   	
\usepackage{geometry}                		
\pdfoutput=1                 		
\usepackage{graphicx}
\usepackage{subfigure}				
\usepackage{amssymb}
\usepackage{url}
\title{\bf Document Image Coding and Clustering for Script Discrimination}
\author{Darko Brodi\'c$^1$\footnote{(Corresponding author)}, Alessia Amelio$^2$, Zoran N. Milivojevi\'c$^3$, Milena Jevti\'c$^1$}
\date{\small $^1$ Technical Faculty in Bor, University of Belgrade, Vojske Jugoslavije 12, 19210 Bor, Serbia, \\\{dbrodic; mjevtic\}@tf.bor.ac.rs\\$^2$ Department of Computer Engineering, Modeling, Electronics and Systems, University of Calabria, \\Via P. Bucci Cube 44, 87036 Rende (CS), Italy,\\aamelio@dimes.unical.it\\$^3$ College of Applied Technical Sciences, Aleksandra Medvedeva 20, Ni\v s 18000, Serbia, \\zoran.milivojevic@vtsnis.edu.rs}							

\begin{document}
\maketitle

\begin{@twocolumnfalse}

{\bf Abstract}.
The paper introduces a new method for discrimination of documents given in different scripts. The document is mapped into a uniformly coded text of numerical values. It is derived from the position of the letters in the text line, based on their typographical characteristics. Each code is considered as a gray level. Accordingly, the coded text determines a 1-D image, on which texture analysis by run-length statistics and local binary pattern is performed. It defines feature vectors representing the script content of the document. A modified clustering approach employed on document feature vector groups documents written in the same script. Experimentation performed on two custom oriented databases of historical documents in old Cyrillic, angular and round Glagolitic as well as Antiqua and Fraktur scripts demonstrates the superiority of the proposed method with respect to well-known methods in the state-of-the-art.
\end{@twocolumnfalse}
\vspace{1cm}

\section{Introduction}
Script recognition has a great importance in document image analysis and optical character recognition \cite{[1]}. Typically, it represents a process of automatic recognition of script by computer in scanned documents \cite{[2]}. This process usually reduces the number of different symbol classes, which is then considered for classification \cite{[3]}.

The proposed methods for script recognition have been classified as global or local ones \cite{[1]}. Global methods divide the image of the document into larger blocks to be normalized and cleaned from the noise. Then, statistical or frequency-domain analysis is employed on the blocks. On the contrary, local methods divide the document image into small blocks of text, called connected components, on which feature analysis, i.e., black pixel runs, is applied \cite{[4]}. This last method is much more computationally heavy than global one, but apt to deal with noisy document images. In any case, previously proposed methods reach an accuracy in script identification between 85\% and 95\% \cite{[1]}.

In this paper, we present a new method for discrimination of documents written in different scripts. In contrast to many previous methods, it can be used prior or during the preprocessing stage. It is primarily based on feature extraction from the bounding box method, its height and center point position in the text line. Hence, there is no need to identify the single characters to differentiate scripts. For this reason, it is particularly useful when the documents are noisy. Furthermore, it maps the connected components of the text to only 4 different codes similarly as in \cite{[5]}, which used character code shapes. In this way, the number of variables is considerably reduced, determining a computer non-intensive procedure. A modified version of a clustering method is proposed and applied to the extracted features for grouping documents given in the same script. Experiments performed on Balkan medieval documents in old Cyrillic, angular and round Glagolitic scripts, and German documents in Antiqua and Fraktur scripts determine an accuracy up to 100\%. The main application of the proposed approach can be used in the cultural heritage area, i.e., in script recognition and classification of historical documents, which includes their origin as well as the influence of different cultural centers to them.

The paper is organized as follows. Section 2 introduces the coding phase and mapping of the text to 1-D image. Section 3 presents the clustering method. Section 4 describes the experiment and discusses it. Finally, Section 5 draws a conclusion.

\section{Script Coding}
Coding phase transforms the script into a uniformly coded text which is subjected to feature extraction. It is composed of two main steps: (i) mapping of the text based on typographical features into an image, by adopting text line segmentation, blob extraction, blob heights and center point detection; (ii) extraction of features from image based on run-length and local binary pattern analysis.

\subsection{Mapping based on typographical features}
\begin{figure}[h]
\begin{center}
\includegraphics[width=13cm,height=13cm,keepaspectratio]{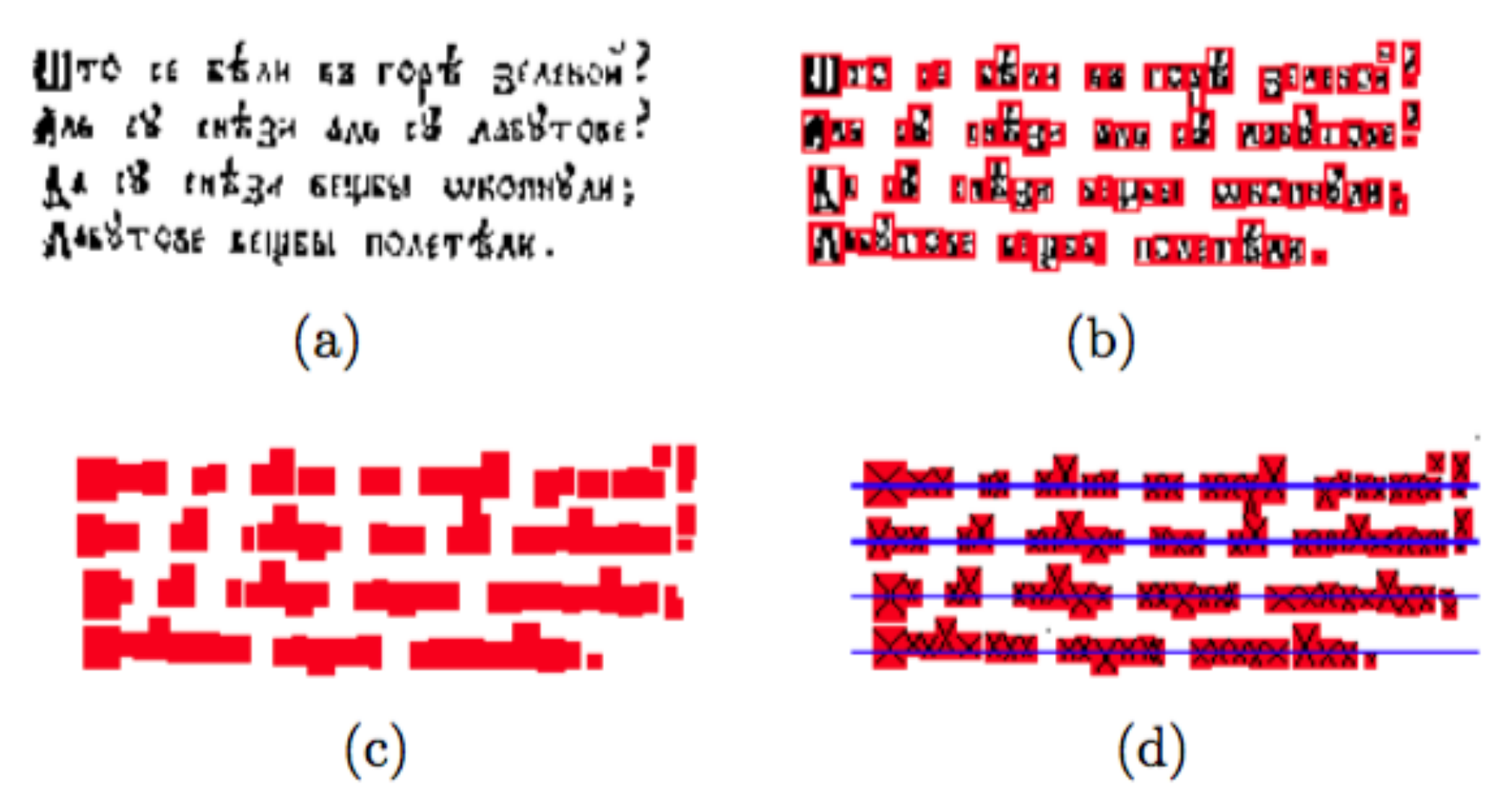}
\caption{(a) Initial text, (b) bounding box detection, (c) bounding box filling, and (d) reference line tracing and center point detection for each bounding box}
\label{fig1}
\end{center}
\end{figure}

First, the text of the document is transformed into a 1-D image based on its typographical features. Text is segmented into text lines by employing the horizontal projection profile. It is adopted for detecting a central line of reference for each text line. A bounding box is traced to each blob, i.e., letter. It is used to derive the distribution of the blob heights and its center point. Typographical classification of the text is based on these extracted features. Figure \ref{fig1} shows this step of the algorithm on a short medieval document from Balkan region written in old Cyrillic script.

\begin{figure}[h]
\begin{center}
\includegraphics[width=14cm,height=14cm,keepaspectratio]{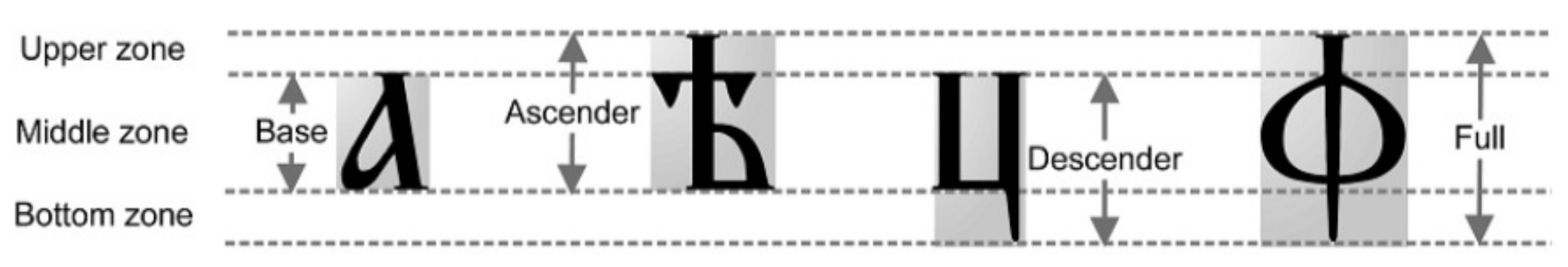}
\caption{Classification of the letters based on typographical features}
\label{fig2}
\end{center}
\end{figure}

Bounding box heights and center point locations can determine the categorization of the corresponding blobs into the following classes \cite{[6]}: (i) base letter (0), (ii) ascender letter (1), (iii) descendent letter (2), and (iv) full letter (3). Figure \ref{fig2} depicts the classification based on typographical features. 

\begin{figure}[h]
\begin{center}
\includegraphics[width=12cm,height=12cm,keepaspectratio]{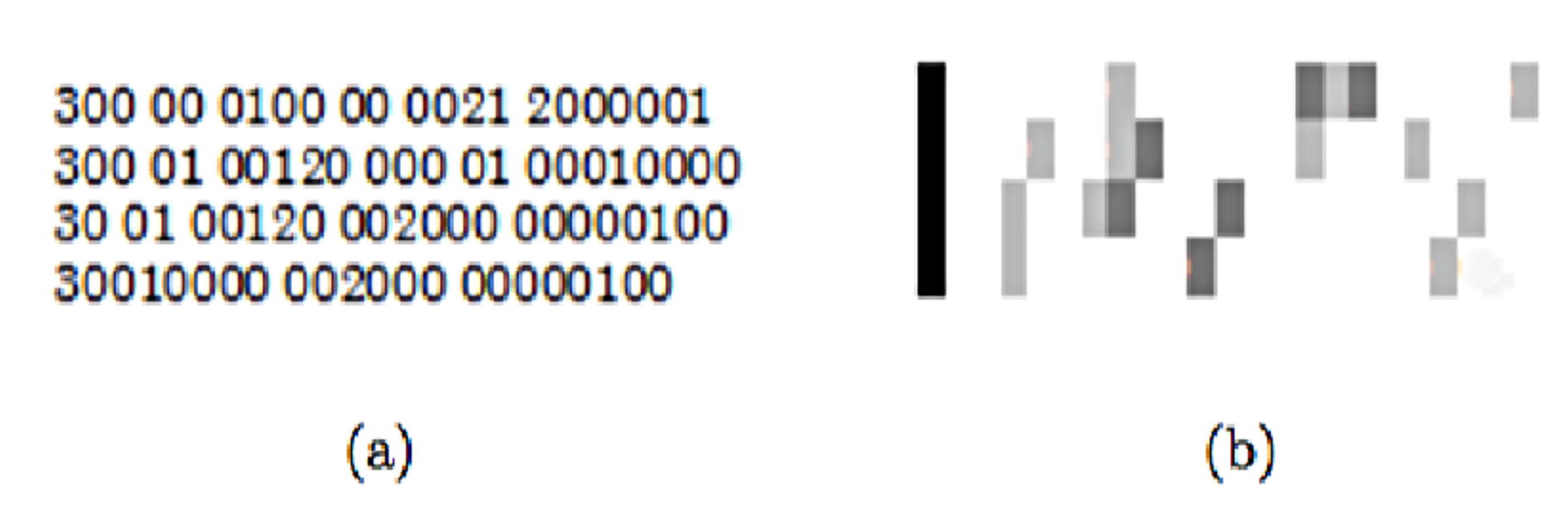}
\caption{(a) Text coding based on typographical features, (b) corresponding image coding}
\label{fig3}
\end{center}
\end{figure}

Starting from this classification, text is transformed into a gray-level 1-D image. In fact, the following mapping is realized: base letter to 0, ascender letter to 1, descendent letter to 2, and full letter to 3 \cite{[7]}. It determines the coding of the text into a long set of numerical codes {0, 1, 2, 3}. Each code has a correspondence with a gray-level, determining the 1-D image. Figure \ref{fig3} shows the procedure of text coding.

\subsection{Feature extraction}
Texture is adopted to compute statistical measures useful to differentiate the images. Run-length analysis can be employed on the obtained 1-D image to create a feature vector of 11 elements representing the document. It computes the following features: (i) short run emphasis (SRE), (ii) long run emphasis (LRE), (iii) gray-level non-uniformity (GLN), (iv) run length non-uniformity (RLN), (v) run percentage (RP) \cite{[8]}, (vi) low gray-level run emphasis (LGRE) and (vii) high gray-level run emphasis (HGRE) \cite{[9]}, (viii) short run low gray-level emphasis (SRLGE), (ix) short run high gray-level emphasis (SRHGE), (x) long run low gray-level emphasis (LRLGE), and (xi) long run high gray-level emphasis (LRHGE) \cite{[10]}. Local Binary Pattern (LBP) analysis can be suitable to obtain only 4 different features from Ô00Õ to Ô11Õ, if the document is represented by 4 gray level images \cite{[11]}. However, this number of features is not sufficient for a good discrimination. Hence, LBP is extended to Adjacent Local Binary Pattern (ALBP) \cite{[12]}, which is the horizontal co-occurrence of LBP. It determines 16 features from Ô0000Õ to Ô1111Õ, from which the histogram is computed as a 16-dimensional feature vector \cite{[13]}. Run-length feature vectors and ALBP feature vectors can be employed for classification and discrimination of scripts in text documents.

\section{Clustering Analysis}
Discrimination of feature vectors representing documents in different scripts is performed by an extension of Genetic Algorithms Image Clustering for Document Analysis (GA-ICDA) method \cite{[14]}. GA-ICDA is a bottom-up evolutionary strategy, for which the document database is represented as a weighted graph $G = (V, E, W )$. Nodes $V$ correspond to documents and edges $E$ to weighted connections, where $W$ is the set of weights, modeling the affinity degree among the nodes. A node $v \in V$ is linked to a subset of its $h$-nearest neighbor nodes $nn^h_v = \{nn^h_v(1), . . . , nn^h_v(k)\}$. They represent the $k$ documents most similar to the document of that node. Similarity is based on the $L_1$ norm of the corresponding feature vectors, while $h$ parameter influences the size of the neighborhood. Hence, the similarity $w(i,j)$ between two documents $i$ and $j$ is expressed as:

\begin{equation}
w(i,j)=e^{-\frac{d(i,j)^2}{a^2}}
\end{equation}

where $d(i,j)$ is the $L_1$ norm between $i$ and $j$ and $a$ is a local scale parameter.

Then, a node ordering $f$ is established, which is a one-to-one association between graph nodes and integer labels, $f: V \to \{1,2,...,n\}$, $n = |V|$. Given the node $v$, the difference is computed between its label $f(v)$ and the labels of the nodes in $nn^h_v$.
Hence, edges are considered only between $v$ and the nodes in $nn^h_v$ for which the label difference $|f(v) - f(nn^h_v(j)|$ is less than a threshold $T$. It is employed for each node in $V$, to realize the adjacency matrix of $G$ with low bandwidth. It represents a graph where the connected components, which are the clusters of documents in a given script, are better visible.

Finally, $G$ is subjected to an evolutionary clustering method to detect clusters of nodes. Then, to refine the obtained solution, a merging procedure is applied on clusters. At each step, the pair of clusters $<C_i,C_j>$ with minimum mutual distance is selected and merged, until a fixed cluster number is reached. The distance between $C_i$ and $C_j$ is computed as the $L_1$ norm between the two farthest document feature vectors, one for each cluster.

A modification is introduced in the base version of GA-ICDA to be more suitable with complex discrimination tasks like differentiation of historical documents given in different scripts. It consists of extending the similarity concept expressed in Equation (1) to a more general characterization. It is realized by substituting the exponent '2' in Equation (1) with a parameter $\alpha$, to obtain a ÒsmoothedÓ similarity computation between the nodes in $G$, when necessary. It is very useful in such a complex context, where documents appear as variegated, for which their mutual distance can be particularly high, even if they belong to the same script typology. Because a lower exponent in Equation (1) determines a higher similarity value from the corresponding distance value, it allows to mitigate the problem.

Hence, the similarity $w(i,j)$ between two documents $i$ and $j$ is now defined as:
\begin{equation}
w(i,j)=e^{-\frac{d(i,j)^\alpha}{a^2}}
\end{equation}

\section{Experimental Results}
The proposed method is evaluated on two complex custom oriented databases. The first one is a collection of labels from Balkan region hand-engraved in stone and hand-printed on paper written in old Cyrillic, angular and round Glagolitic scripts. The database contains 5 labels in old Cyrillic, 10 labels in angular and 5 labels in round Glagolitic, for a total of 20 labels. The second database is composed of 100 historical German documents mainly from the J. W. von GoetheÕs poems, written in Antiqua and Fraktur scripts. The experiment consists of employing the modified GA-ICDA on the run-length and ALBP feature vectors computed from the documents in the two databases, for testing the efficacy in correctly differentiating the script types. A comparison is performed between GA-ICDA with modification and other 4 clustering methods: the base version of GA-ICDA, Complete Linkage Hierarchical clustering, Self-Organizing-Map (SOM) and K-Means, well-known for document categorization \cite{[15]}. A trial and error procedure is applied on benchmark documents, different from the databases, for tuning the parameters of the methods. Those providing the best solution on the benchmark are employed for clustering. Hence, $\alpha$ parameter is fixed to 1. Precision, Recall, F-Measure (computed for each script class) and Normalized Mutual Information (NMI) are adopted as performance measures for clustering evaluation \cite{[16]}. Each method has been executed 100 times and average value of measures together with standard deviation have been computed.

Tables \ref{tab1} and \ref{tab2} report the results of the experiment respectively on the first and second database. 

\begin{table}[!h]
\caption{Clustering results on the first database. Standard deviation is reported in parenthesis.}
\begin{center}
\includegraphics[width=11.5cm,height=11.5cm,keepaspectratio]{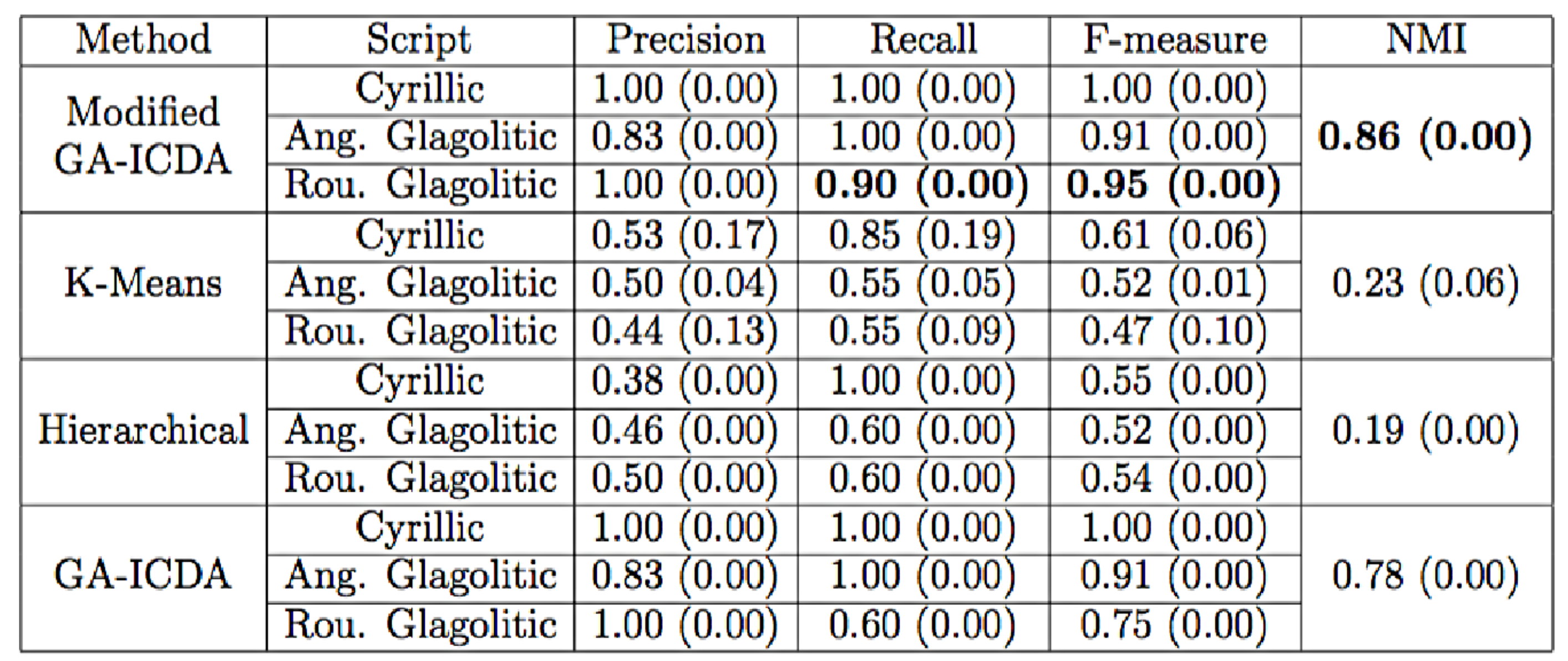}
\end{center}
\label{tab1}
\end{table}%

\begin{figure}
\begin{center}
\includegraphics[width=15.5cm,height=15.5cm,keepaspectratio]{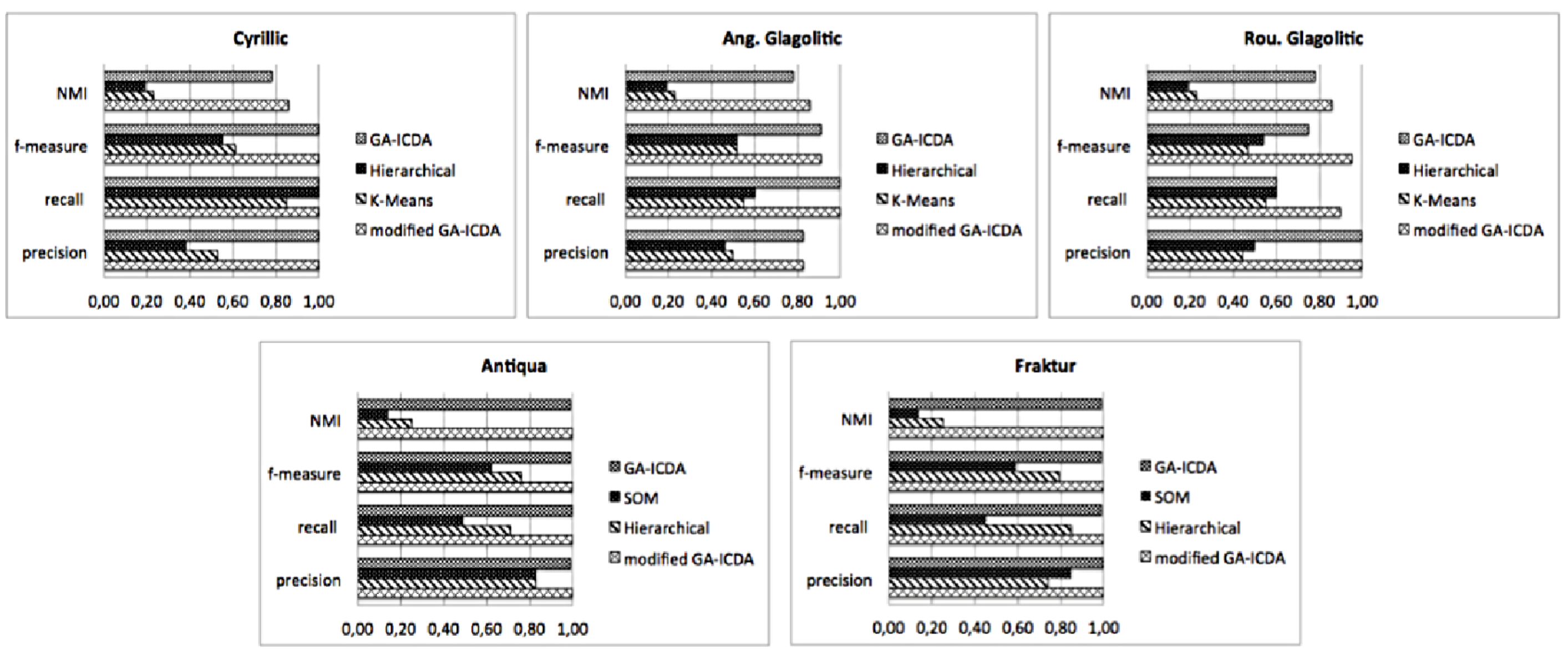}
\caption{Results of the experiment on the first (top) and second (bottom) database}
\label{fig4}
\end{center}
\end{figure}

Figure \ref{fig4} shows the corresponding results in graphical form. It is worth noting that GA-ICDA with modification performs considerably better than the other clustering methods for both the databases and that adopted modification determines an improvement in the final result with respect to the base version of GA-ICDA. Also, the standard deviation is always zero. It confirms the stability of the obtained results.

\begin{table}
\caption{Clustering results on the second database. Standard deviation is given in parenthesis.}
\begin{center}
\includegraphics[width=10cm,height=10cm,keepaspectratio]{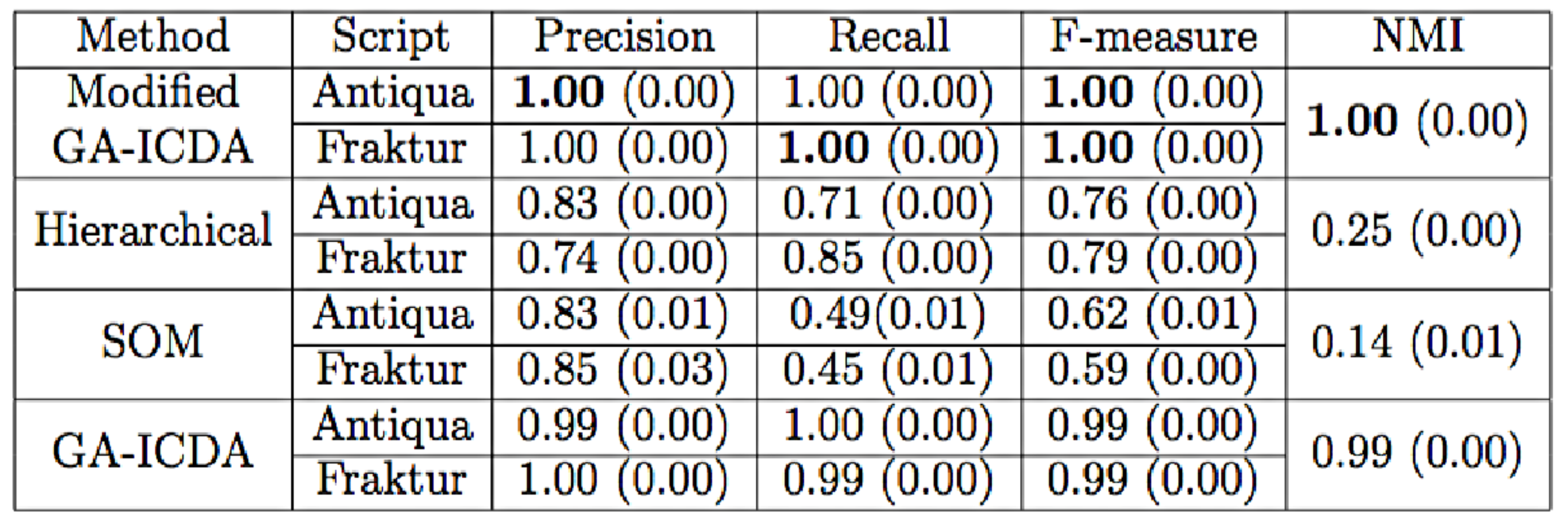}
\end{center}
\label{tab2}
\end{table}%

\section{Conclusions}
The paper proposed a new method for differentiation of script type in text documents. In the first step, the document was mapped into a uniformly coded text. Then, it was transformed into 1-D gray-level image, from which texture features were extracted. A modified version of the GA-ICDA method was adopted on feature vectors for document discrimination based on script typology. A huge experimentation on two complex databases of historical documents proved the effectiveness of the proposed method. 

Future work will extend the experiment on large datasets of labels engraved on different materials, like bronze, and will compare the method with other classification algorithms.

\end{document}